\documentclass[10pt,twocolumn,letterpaper]{article}

\usepackage{iccv}
\usepackage{times}
\usepackage{epsfig}
\usepackage{graphicx}
\usepackage{amsmath}
\usepackage{amssymb}
\usepackage{bbm}
\usepackage{mathtools}
\usepackage{booktabs}
\usepackage{xcolor,colortbl}
\usepackage{algorithm,algorithmic}
\usepackage{caption, subcaption}
\usepackage{comment}
\usepackage{multirow}
\usepackage{authblk}
\usepackage{bm}
\usepackage[accsupp]{axessibility}  %

\usepackage{svg}

\usepackage[pagebackref=true,breaklinks=true,letterpaper=true,colorlinks,bookmarks=false]{hyperref}

\definecolor{Gray}{gray}{0.9}

\iccvfinalcopy %

\def\iccvPaperID{10230} %

\ificcvfinal\fi

\begin{document}

\title{Poincar\'e ResNet}

\author[1]{Max van Spengler}
\author[2]{Erwin Berkhout}
\author[1]{Pascal Mettes}

\affil[1]{
    VIS Lab\\
    Informatics Institute\\
    University of Amsterdam\\
}
\affil[2]{
    Department of Oral Radiology\\
    Academic Center for Dentistry\\
    University of Amsterdam \& VU Amsterdam
}

\maketitle
\ificcvfinal\thispagestyle{empty}\fi

\begin{abstract}
This paper introduces an end-to-end residual network that operates entirely on the Poincar\'e ball model of hyperbolic space. Hyperbolic learning has recently shown great potential for visual understanding, but is currently only performed in the penultimate layer(s) of deep networks. All visual representations are still learned through standard Euclidean networks. In this paper we investigate how to learn hyperbolic representations of visual data directly from the pixel-level. We propose Poincar\'e ResNet, a hyperbolic counterpart of the celebrated residual network, starting from Poincar\'e 2D convolutions up to Poincar\'e residual connections. We identify three roadblocks for training convolutional networks entirely in hyperbolic space and propose a solution for each: (i) Current hyperbolic network initializations collapse to the origin, limiting their applicability in deeper networks. We provide an identity-based initialization that preserves norms over many layers. (ii) Residual networks rely heavily on batch normalization, which comes with expensive Fr\'echet mean calculations in hyperbolic space. We introduce Poincar\'e midpoint batch normalization as a faster and equally effective alternative. (iii) Due to the many intermediate operations in Poincar\'e layers,
the computation graphs of deep learning libraries blow up, limiting our ability to train on deep hyperbolic networks. We provide manual backward derivations of core hyperbolic operations to maintain manageable computation graphs.
\end{abstract}

\section{Introduction}
Deep learning in hyperbolic space has gained traction in recent years empowered by their inherent ability to embed hierarchical data with arbitrarily low distortion \cite{sarkar2011trees} and by being more compact and dense \cite{chami2019, nickel2017poincare, shimizu2021}. These promising characteristics have led to rapid developments in hyperbolic representation learning for tree-like structures \cite{balazevic2019multi,chami2020trees,ganea2018entailment,law2019lorentzian,nickel2017poincare,sala2018representation}, graphs \cite{chami2019,dai2021hyperbolic,liu2019hyperbolic,zhang2021hyperbolic}, text \cite{chen2021fully,dhingra2018embedding,tifrea2019poincar}, action skeletons \cite{franco2023hyperbolic}, biological structures \cite{klimovskaia2020poincare}, and more.

Recently, hyperbolic learning has also been investigated for visual understanding. Hyperbolic embeddings of images and videos have been shown to improve few-shot learning \cite{fang2021kernel,gao2021curvature,guo2022clipped,ma2022adaptive,zhang2022hyperbolic}, hierarchical recognition \cite{dhall2020hierarchical,ghadimi2021hyperbolic,long2020searching,long2023cross,yu2022skin}, segmentation \cite{chen2022hyperbolic,atigh2022hyperbolic} and metric learning \cite{ermolov2022hyperbolic,zhang2021learning} amongst others. While promising, the use of hyperbolic geometry in computer vision has been limited to the classifier space, with visual representations being learned on conventional networks that operate in Euclidean space.

This paper explores
learning visual representations entirely in hyperbolic space. The ability to learn hyperbolic representations directly from the pixel-level will allow us to unlock the broad potential of hyperbolic geometry for vision, such as capturing latent hierarchical visual representations \cite{khrulkov2020hyperbolic}, training compact network architectures \cite{chami2019, nickel2017poincare, shimizu2021}, and creating networks that better mimic visual representation learning in the brain \cite{zhang2022hippocampal}. Empowered by successful implementations of non-visual layers~\cite{ganea2018,shimizu2021}, the time is ripe for visual hyperbolic feature learning.

As a step towards fully hyperbolic visual learning, we start from the highly celebrated ResNet \cite{he2015b} and rebuild its architecture in hyperbolic space; from 2D convolutions to residual connections. Optimizing a ResNet in the Poincar\'e ball model comes with several challenges. First, we find that existing network initializations in hyperbolic space lead to vanishing signals, which derail learning over many convolutional layers. We provide an identity-based network initialization that preserves the output norm over many layers. Second, ResNets rely extensively on batch normalization, but its generalization to hyperbolic space requires expensive Fr\'echet mean calculations \cite{lou2020}. We introduce Poincar\'e midpoint batch normalization, which allows us to compute approximate means at a fraction of the computational cost. Third, the basic gyrovector operations in the Poincar\'e ball model consist of many intermediate calculations. In modern deep learning libraries, all these calculations are stored for automatic differentiation, blowing up the computation graph. We have derived and implemented the backward pass of core hyperbolic gyrovector operations to contain the computation graph.

Empirically, we show that our network initialization is indeed norm-preserving and improves network generalization. We show that our midpoint batch normalization speeds up training by 25\% with no loss in classification accuracy. We furthermore demonstrate the potential of Poincar\'e ResNet for out-of-distribution detection, adversarial robustness, and learning complementary representations compared to Euclidean ResNet. {\color{black}The code is available at \url{https://github.com/maxvanspengler/poincare-resnet} with a similar implementation in the documentation of HypLL \cite{spengler2023hypll}.}

\section{Background and related work}
\subsection{Poincar\'e ball model of hyperbolic space}
This paper operates on the most commonly used model of hyperbolic geometry in deep learning, namely the Poincar\'e ball model. We will therefore restrict the background discussion to this model and refer to Peng \etal~\cite{peng2021} for a more comprehensive discussion on the different isometric models of hyperbolic space.
The $n$-dimensional Poincar\'e ball model with constant negative curvature $-c$ is defined as the Riemannian manifold $(\mathbb{B}_c^n, \mathfrak{g}_c)$, where 
\begin{equation}
    \mathbb{B}_c^n = \{\bm{x} \in \mathbb{R}^n : ||\bm{x}||^2 < \frac{1}{c}\},
\end{equation}
and where
\begin{equation}
    \mathfrak{g}_c = \lambda_{\bm{x}}^c I_n, \quad \lambda_{\bm{x}}^c = \frac{2}{1 - c ||\bm{x}||^2},
\end{equation}
with $I_n$ being the $n$-dimensional identity matrix. 
The Poincar\'e ball model can be turned into a gyrovector space \cite{ungar2009} by endowing it with M\"obius addition and M\"obius scalar multiplication, respectively defined as
\begin{equation}
\begin{split}
    \bm{x} \oplus_c \bm{y} = & \frac{(1 + 2c \langle \bm{x}, \bm{y} \rangle + c ||\bm{y}||^2) \bm{x} + (1 - c ||\bm{x}||^2) \bm{y}}{1 + 2c \langle \bm{x}, \bm{y} \rangle + c^2 ||\bm{x}||^2 ||\bm{y}||^2},\\
    r \otimes_c \bm{x} = &\frac{1}{\sqrt{c}} \tanh \big(r \tanh^{-1} (\sqrt{c} ||\bm{x}||)\big) \frac{\bm{x}}{||\bm{x}||},
\end{split}
\end{equation}
where $\bm{x}, \bm{y} \in \mathbb{B}_c^n$, $r \in \mathbb{R}$ and where $||\cdot||$ and $\langle \cdot, \cdot \rangle$ denote the Euclidean norm and inner product, respectively. An important map related to this gyrovector space is the gyrator $\text{gyr} : \mathbb{B}_c^n \times \mathbb{B}_c^n \rightarrow \text{Aut}(\mathbb{B}_c^n, \oplus_c)$, where $\text{Aut}(\mathbb{B}_c^n, \oplus_c)$ denotes the set of automorphisms on $\mathbb{B}_c^n$ \cite{ungar2009}. This map is implicitly defined as
\begin{equation}
    \text{gyr}[\bm{x}, \bm{y}] \bm{z} = - (\bm{x} \oplus_c \bm{y}) \oplus_c \big(\bm{x} \oplus_c (\bm{y} \oplus_c \bm{z})\big),
\end{equation}
where $\bm{x}, \bm{y}, \bm{z} \in \mathbb{B}_c^n$, which can be used to measure the extent to which M\"obius addition deviates from commutativity. It will be used later on to define parallel transport. Furthermore, we can compute the distance between any two points $\bm{x}, \bm{y} \in \mathbb{B}_c^n$ as
\begin{equation}
    d_c(\bm{x}, \bm{y}) = \frac{2}{\sqrt{c}} \tanh^{-1} (\sqrt{c} ||-\bm{x} \oplus_c \bm{y}||).
\end{equation}
For an in-depth analysis of this gyrovector space approach to the Poincar\'e ball see \cite{ungar2009}.
Using the definition of M\"obius addition, the exponential and logarithmic maps can be written as \cite{ganea2018}
\begin{equation*}
\begin{split}
    \exp_{\bm{x}}^c (\bm{v}) &= \bm{x} \oplus_c \Big(\tanh\big(\frac{\sqrt{c} \lambda_{\bm{x}}^c ||\bm{v}||}{2}\big) \frac{\bm{v}}{\sqrt{c} ||\bm{v}||}\Big),\\
    \log_{\bm{x}}^c (\bm{y}) &= \frac{2}{\sqrt{c} \lambda_{\bm{x}}^c} \tanh^{-1} \big(\sqrt{c} ||-\bm{x} \oplus_c \bm{y}||\big) \frac{-\bm{x} \oplus_c \bm{y}}{||-\bm{x} \oplus_c \bm{y}||},
\end{split}
\end{equation*}
where $\bm{x}, \bm{y} \in \mathbb{B}_c^n$ and $\bm{v} \in \mathcal{T}_{\bm{x}} \mathbb{B}_c^n$. Moreover, we can define parallel transport $P_{\bm{x} \rightarrow \bm{y}}^c : \mathcal{T}_{\bm{x}} \mathbb{B}_c^n \rightarrow \mathcal{T}_{\bm{y}} \mathbb{B}_c^n$ as follows \cite{shimizu2021}
\begin{equation}
    P_{\bm{x} \rightarrow \bm{y}}^c (\bm{v}) = \frac{\lambda_{\bm{x}}^c}{\lambda_{\bm{y}}^c} \text{gyr}[\bm{y}, -\bm{x}]\bm{v},
\end{equation}
which allows us to transport a tangent vector at a point $\bm{x} \in \mathbb{B}_c^n$ to the tangent space at another point $\bm{y} \in \mathbb{B}_c^n$, used for example in batch normalization.

\subsection{The Poincar\'e ball model in neural networks}
\label{subsect:poincare_nn}
To perform deep learning on the Poincar\'e ball model, Ganea \etal~\cite{ganea2018} outline a theoretical framework for incorporating this model into core layers of neural networks, such as hyperbolic logistic regression, hyperbolic fully-connected, and hyperbolic recurrent layers.
More recently, Shimizu \etal~\cite{shimizu2021} made important improvements to this framework to ensure that the hyperbolic geometry was fully taken advantage of without the need for additional learnable parameters. We will therefore use this work as a starting point for the rest of this paper and provide a short overview here.

As a foundation, Poincar\'e multinomial logistic regression is defined by computing the score for each of $n$ classes for some input $\bm{x} \in \mathbb{B}_c^m$ as
\begin{align*}
    v_k (\bm{x}) = \frac{2}{\sqrt{c}} ||\bm{z}_k|| \sinh^{-1} \Big(\lambda_{\bm{x}}^c \langle \sqrt{c} \bm{x}, \frac{\bm{z}_k}{||\bm{z}_k||} \rangle \cosh(2 \sqrt{c} r_k) \\
    - (\lambda_{\bm{x}}^c - 1) \sinh(2 \sqrt{c} r_k)\Big),
\end{align*}
where $\bm{z}_k \in \mathcal{T}_{\bm{0}} \mathbb{B}_c^m = \mathbb{R}^m$ and $r_k \in \mathbb{R}$ are the parameters for the $k$-th class. These scores are equivalent to the distances between the input $\bm{x}$ and the $n$ different Poincar\'e hyperplanes determined by the parameters $\{(\bm{z}_k, r_k)\}_{i=1}^n$. Here, $\bm{z}_k$ determines the orientation of the hyperplane while $r_k$ determines its offset with respect to the origin. A Poincar\'e fully connected layer mapping input $\bm{x} \in \mathbb{B}_c^m$ to $\mathbb{B}_c^n$ is in turn defined as
\begin{equation}\label{eq:FC_layer}
    \bm{y} = \mathcal{F}^c (\bm{x}; Z, \bm{r}) = \frac{\bm{w}}{1 + \sqrt{1 + c ||\bm{w}||^2}},
\end{equation}
with 
\begin{equation}
    \bm{w} = \Big(\frac{1}{\sqrt{c}} \sinh(\sqrt{c} v_k (\bm{x}))\Big)_{k=1}^n,
\end{equation}
where the $v_k (\cdot)$ are the scores from the Poincar\'e multinomial logistic regression and where $Z = [\bm{z}_1 | \ldots | \bm{z}_n] \in (\mathcal{T}_{\bm{0}} \mathbb{B}_c^m)^n = \mathbb{R}^{m \times n}$ and $\bm{r} = (r_k)_{k=1}^n \in \mathbb{R}^m$ are the parameters of the layer. Given hyperbolic fully connected layers, Shimizu \etal~\cite{shimizu2021} outline general formulations for self-attention and convolutional operations in hyperbolic space. We take such investigations to the visual domain and arrive at Poincar\'e ResNets, which require 2D convolutions, fast batch normalization, residual blocks, norm-preserving initialization and derived backpropagation of core operations in order to be realized.

\subsection{Hyperbolic learning in computer vision}
Khrulkov \etal~\cite{khrulkov2020hyperbolic} have shown that both image data and labels contain hierarchical structures and introduced Hyperbolic Image Embeddings to exploit these observations. In their approach, embeddings of images from standard networks are mapped to hyperbolic space, followed by a final classification layer based on hyperbolic logistic regression or hyperbolic prototypical learning, directly improving few-shot learning and uncertainty quantification.

A wide range of works have investigated hyperbolic visual embeddings, see Mettes \etal \cite{mettes2023hyperbolic}. Several works have proposed prototypes-based hyperbolic embeddings for few-shot learning \cite{fang2021kernel,gao2021curvature,guo2022clipped,ma2022adaptive,zhang2022hyperbolic}, where hyperbolic space consistently outperforms Euclidean space. Hyperbolic embeddings of classes based on their hierarchical relations has also shown to be effective for zero-shot learning \cite{liu2020hyperbolic,xu2022meta} and hierarchical recognition \cite{dhall2020hierarchical,ghadimi2021hyperbolic,long2020searching,yu2022skin}. Hyperbolic embeddings have furthermore been effective in metric learning \cite{ermolov2022hyperbolic,zhang2021learning}, object detection \cite{valada2022hyperbolic}, image segmentation \cite{chen2022hyperbolic,atigh2022hyperbolic} and future prediction in videos~\cite{suris2021learning}. 

In generative learning, hyperbolic variational auto-encoders \cite{hsu2021capturing,mathieu2019continuous,nagano2019wrapped}, generative adversarial networks \cite{lazcano2021hgan} and normalizing flows \cite{bose2020latent,mathieu2020riemannian} have been shown to obtain competitive results in data-constrained settings. A number of recent works have proposed unsupervised hyperbolic learning approaches \cite{hsu2021capturing,monath2019gradient,weng2021unsupervised,yan2021unsupervised}, allowing for learning and discovering hierarchical representations.

This body of literature highlights that hyperbolic geometry is fruitful for visual understanding. In current literature, however, hyperbolic learning is restricted to the final embedding layers, with all visual representations being learned by standard networks. This paper strives to learn hyperbolic representations in an end-to-end manner, from pixels to labels, complementing current research on computer vision with hyperbolic embeddings.

\section{Poincar\'e residual networks for images}\label{sect:method}
We consider the problem of image classification where our dataset is denoted by $(\bm{x}_i, y_i)_{i=1}^N$, with $\bm{x}_i \in \mathbb{R}^{H \times W \times 3}$ and $y_i \in \{1, \ldots, C\}$. Here, $\bm{x}_i$ denotes the pixel values of the $i$-th input image with height $H$ and width $W$, while $y_i$ denotes the corresponding label. Our goal is to train a network $y = \phi(\bm{x})$ that maps an input image $\bm{x}$ to a label $y$. Specifically, we strive to formulate the celebrated ResNet \cite{he2015b} architecture in the Poincar\'e ball model.

In residual networks, the basic building block consists of two weight layers with a ReLU activation between the layers. Afterwards, the input is added to the transformed output through a residual connection, followed by another ReLU activation. A weight layer is typically given as a convolutional layer followed by a batch normalization.
Thus, to create Poincar\'e residual blocks, all these operations need to be formulated in hyperbolic space. Below, we separately outline how to formalize and construct (i) Poincar\'e 2D convolutions and residual blocks, (ii) how to initialize hyperbolic networks, (iii) Poincar\'e midpoint batch normalization, and (iv) forward and backward propagation of core hyperbolic operations.

\subsection{Poincar\'e convolutions and residuals}
We start by formalizing 2D convolutional operations for images in the Poincar\'e ball model using the approach of Shimizu \etal~\cite{shimizu2021}. Suppose we have an input image $\bm{x}$ with pixel values
\begin{equation}
    \bm{x}_{ij} \in \mathbb{B}_{c}^{C_{in}}, \quad i = 1, \ldots, H_{in}, \enspace j = 1, \ldots, W_{in},
\end{equation}
where $C_{in}$ is the number of input channels and where $H_{in}$ and $W_{in}$ are the height and width of the image, respectively. Then we can define a 2D Poincar\'e convolution operation with $C_{out}$ output channels and with receptive field size $K \times K$, with $K$ odd.
This approach and its Euclidean counterpart have the same grid connections between the input values and output values. Only the convolutional operations behind these connections are defined differently. So, the output will have pixel values
\begin{equation}
    \bm{h}_{ij} \in \mathbb{B}_c^{C_{out}}, \quad i = 1, \ldots, H_{out}, \enspace j = 1, \ldots, W_{out},
\end{equation}
where $\bm{h}_{kl}$ is computed from the pixels $\bm{x}_{ij}$ in the receptive field at that position, so where
\begin{align}
\begin{split}
    k - \Big\lfloor\frac{K}{2}\Big\rfloor \leq i \leq k + \Big\lfloor \frac{K}{2} \Big\rfloor, \\
    l - \Big\lfloor \frac{K}{2}\Big\rfloor \leq j \leq l + \Big\lfloor \frac{K}{2} \Big\rfloor.
\end{split}
\end{align}
We denote this receptive field at position $(k, l)$ by $X_{kl}$. Note that $H_{out}$ and $W_{out}$ depend on the input dimensions, the receptive field size $K$ and, optionally, on stride and padding.

\begin{figure}
    \centering
    \includegraphics[width=8.5cm]{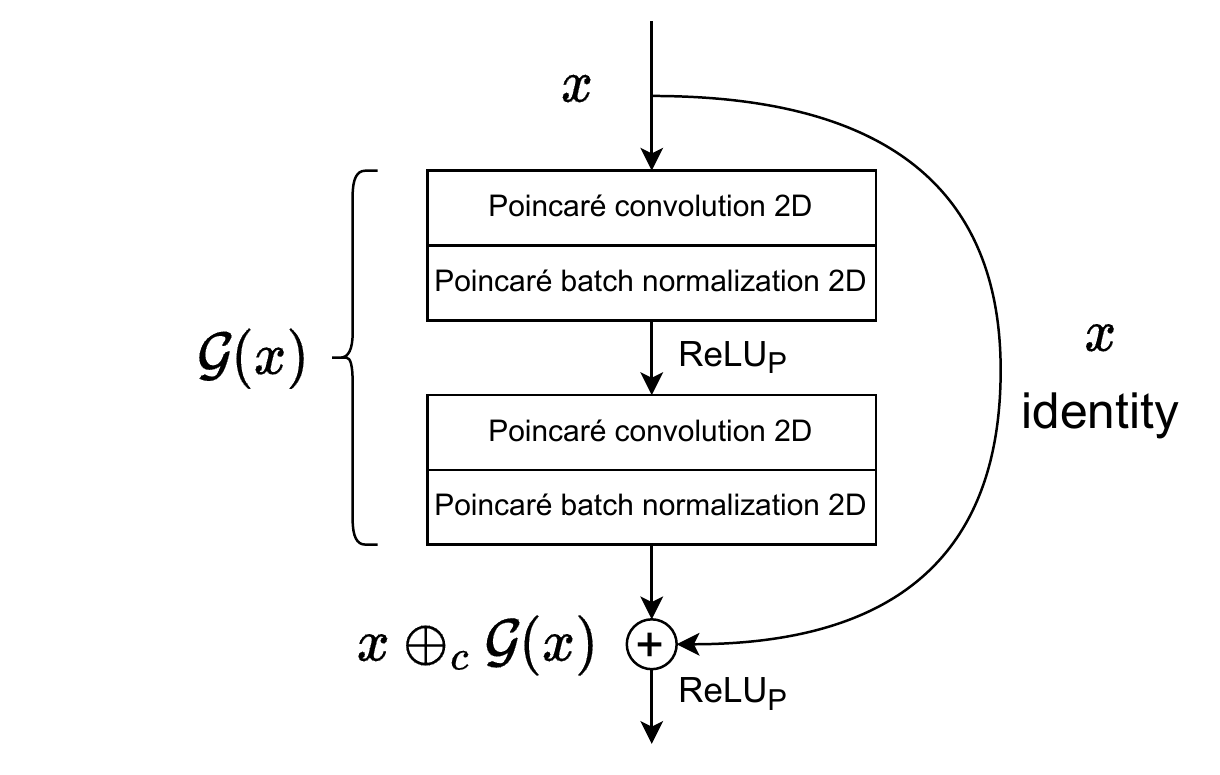}
    \caption{\textbf{A Poincar\'e residual block}, the basic building block of our Poincar\'e ResNet architectures and a direct generalization of the original residual block of He \etal \cite{he2015b}.}
    \label{fig:residual_block}
\end{figure}

Similar to the Euclidean convolutional layer, for each $\bm{h}_{kl}$, we want to apply a fully connected layer to the concatenation of the vectors within the receptive field, so we want to compute the output as
\begin{equation}
    \bm{h}_{kl} = \mathcal{F}^c ({\parallel} X_{kl}; Z, \bm{r}),
\end{equation}
where ${\parallel} \cdot$ denotes some concatenation operation and $\mathcal{F}^c$ is the Poincar\'e fully-connected layer defined in equation (\ref{eq:FC_layer}) with parameters $Z$ and $\bm{r}$. Note that the usual concatenation is inappropriate for vectors on the Poincar\'e ball as this can result in vectors outside the manifold. We therefore employ $\beta$-concatenation as an alternative, which is a concatenation operation that preserves the expectation of the Poincar\'e norm of the output vector \cite{shimizu2021}. This operation, applied to $M$ Poincar\'e vectors $\{\bm{b}_i \in \mathbb{B}_c^{n_i}\}_{i=1}^M$ with $n = \sum_i n_i$, is defined in three steps:
\begin{enumerate}
    \item Map each of the vectors to the tangent space at the origin of their respective Poincar\'e balls: $\bm{v}_i = \log_0^c (\bm{b}_i)$;
    \item Let $\beta_n = B(\frac{n}{2}, \frac{1}{2})$, with $B$ the beta function, scale each of the vectors $\bm{v}_i$ by $\beta_n \beta_{n_i}^{-1}$ and let $\bm{v}$ be the concatenation of these scaled vectors, so $\bm{v} = (\beta_n \beta_{n_1}^{-1} \bm{v}_1^T, \ldots, \beta_n \beta_{n_N}^{-1} \bm{v}_N^T)^T$;
    \item Project the resulting vector back onto the $n$-dimensional Poincar\'e ball: $\exp_0^c (\bm{v})$.
\end{enumerate}
We denote this operation by $\prescript{}{\beta}{\parallel} \cdot$. Now, we can write the 2D Poincar\'e convolution operation as
\begin{equation}
    \bm{h}_{kl} = \mathcal{F}^c (\prescript{}{\beta}{\parallel} X_{kl}; Z, \bm{r}),
\end{equation}
where $k = 1, \ldots, H_{out}$, $l = 1, \ldots, W_{out}$ and where $\mathcal{F}^c$ maps from $\mathbb{B}_c^{K^2 \times C_{in}}$ to $\mathbb{B}_c^{C_{out}}$. 

Next, we define a Poincar\'e version of the residual block by replacing the convolutional layers by Poincar\'e convolutional layers and by applying a hyperbolic batch normalization, which will be defined in the next subsection. Pointwise nonlinearities can still be applied in the tangent space at the origin of the Poincar\'e ball by using the logarithmic and exponential maps. So, the Poincar\'e version of the ReLU nonlinearity becomes
\begin{equation}
    \text{ReLU}_P = \exp_0^c \circ \; \text{ReLU} \circ \log_0^c,
\end{equation}
where $\circ$ denotes function composition. We will use this Poincar\'e version to replace the two ReLU nonlinearities.
We can furthermore replace the skip connection by $\bm{x} \oplus_c \mathcal{G} (\bm{x})$, where $\mathcal{G}$ denotes the transformation given by the two Poincar\'e convolutional layers and Poincar\'e batch normalizations. Figure \ref{fig:residual_block} visualizes the Poincar\'e residual block.

\begin{algorithm}[tb]
    \caption{Poincar\'e midpoint batch normalization}
    \label{alg:batch_norm}
    \begin{algorithmic}
        \STATE {\bfseries Training Input:} Data batches $\{\bm{x}_1^{(t)}, \ldots, \bm{x}_m^{(t)}\} \subseteq \mathbb{B}_c^n$ for $t \in [1, \ldots, T]$, testing momentum $\eta \in [0, 1]$
        \STATE {\bfseries Learned Parameters:} $\beta \in \mathbb{B}_c^n$, $\gamma \in \mathbb{R}$
        \STATE {\bfseries Normalization Algorithm:} 
        \FOR{$t = 1, \ldots, T$}
            \STATE $\mu \gets \text{Poincar\'eMidpoint}(\{\bm{x}_1^{(t)}, \ldots, \bm{x}_m^{(t)}\})$
            \STATE $\sigma^2 \gets \frac{1}{m} \sum_{i=1}^m d(\bm{x}_i^{(t)}, \mu)^2$
            \FOR{$i = 1, \ldots, m$}
                \STATE $\tilde{\bm{x}}_i^{(t)} \gets \exp_{\beta}^c \Big( \sqrt{\frac{\gamma}{\sigma^2}} P_{\mu \rightarrow \beta}^c (\log_{\mu}^c \bm{x}_i^{(t)}) \Big)$
            \ENDFOR
            \RETURN normalized batch $\tilde{\bm{x}}_1^{(t)}, \ldots, \tilde{\bm{x}}_m^{(t)}$
        \ENDFOR
    \end{algorithmic}
\end{algorithm}

\subsection{Poincar\'e midpoint batch normalization}
In a residual block, each convolutional layer is immediately followed by a batch normalization step. Lou \etal \cite{lou2020} have previously defined a Poincar\'e version of batch normalization based on their iterative approximation to the Fr\'echet mean.
While more efficient than previously available methods, this iterative approach still makes the Fr\'echet mean a computationally expensive step. Directly plugging a Fr\'echet-based batch normalization in our Poincar\'e ResNet would account for roughly 77\% of the computation in a forward step. We therefore seek to perform batch normalization with greater computational efficiency.

We suggest to take an alternative aggregation of Poincar\'e vectors, namely the Poincar\'e midpoint \cite{ungar2009}. The midpoint of the Poincar\'e vectors $\{\bm{x}_i \in \mathbb{B}_c^n\}_{i=1}^N$ is computed as
\begin{equation}
    \mu = \frac{1}{2} \otimes_c \frac{\sum_{i=1}^N \lambda_{\bm{x}_i}^c \bm{x}_i}{\sum_{i=1}^N (\lambda_{\bm{x}_i}^c - 1)}.
\end{equation}
The resulting midpoint batch normalization algorithm is outlined in Algorithm \ref{alg:batch_norm}.
The goal of batch normalization is to keep feature vectors centered around the origin and to keep the variance of their norms within a manageable range. By replacing the Fr\'echet mean by the Poincar\'e midpoint, the vectors will no longer be centered exactly at the origin, but still close enough to achieve the improved stability that batch normalization normally results in. Moreover, the Poincar\'e midpoint can be computed directly without any iterative methods, making it substantially faster to compute than the Fr\'echet mean.

\subsection{Hyperbolic network initialization} %
The canonical ResNet architecture uses Kaiming initialization, which aims to prevent reduction or magnification of input signals as this would hinder convergence during training \cite{he2015a}. This is achieved by maintaining the variance of the components of both the features and the gradients throughout the network. However, such an approach is inappropriate for the Poincar\'e fully connected and convolutional layers as the components of a Poincar\'e vector are necessarily dependent, since the Euclidean norm of such vectors is bounded by $c^{-\frac{1}{2}}$. 

To that end, Shimizu \etal \cite{shimizu2021} propose to initialize the weights $Z$ of the Poincar\'e fully connected layer through sampling from the normal distribution $\mathcal{N}(0, (2mn)^{-1})$, where $m$ is the input dimension and $n$ the output dimension of the layer. The biases $r$ are initialized as zeros. We find that this initialization results in vanishing signals, where the norm of an input converges to zero after a few layers. To obtain a norm-preserving network initialization in hyperbolic space,
we take the initialization for the weights of a Poincar\'e layer mapping from $\mathbb{B}_{c}^m$ to $\mathbb{B}_{c}^n$ with $m \leq n$ as
\begin{equation}\label{eq:id_init}
    Z = 
    \begin{cases}
    \frac{1}{2} I_n & m = n, \\
    \frac{1}{2} [I_m | O_{m, n-m}] & m < n,
    \end{cases}
\end{equation}
where $I_n$ is the $n \times n$-identity matrix and where $O_{i,j}$ is the $i \times j$-zero matrix. We initialize the biases $r$ as a vector of zeros.
Using this initialization, for $m = n$, we see that
\begin{equation}
    v_k(\bm{x}) = \frac{1}{\sqrt{c}} \sinh^{-1} \Big(\sqrt{c} \lambda_{\bm{x}}^c x_k\Big),
\end{equation}
and, therefore,
\begin{align}
    w = \lambda_{\bm{x}}^c \bm{x},
\end{align}
from which it follows that $\bm{y} = \bm{x}$. When $m < n$, we get $\bm{y} = (\bm{x}^T | \bm{0}_{n-m}^T)^T$ instead, where $\bm{0}_{n-m}$ is an $(n-m)$-dimensional vector of zeros. Thus, for the cases $m \leq n$, this initialization keeps the norms of the vectors constant throughout the network.

For residual networks, $m \leq n$ for each layer except for the linear layer at the end of the network. Therefore, we initialize each of the convolutional layers using our identity initialization. The final linear layer will be initialized using the initialization by \cite{shimizu2021}. We find the vanishing effect of this single layer to be harmless to the performance.

\subsection{Optimization and backward propagation}
For neural networks on Riemannian manifolds, one generally has to consider the manifold on which the parameters live for optimization \cite{bonnabel2013riemanniansgd}. For Poincar\'e residual networks, we need to consider the weights of three different layers, namely, the fully-connected layer, the convolutional layer, and the batch normalization. The parameters of the fully-connected layer and the convolutional layer as proposed by Shimizu \etal~ \cite{shimizu2021} live in Euclidean space, so we can use Euclidean optimizers for these layers. However, the batch normalization algorithm shown in Algorithm \ref{alg:batch_norm} makes use of a parameter vector living on $\mathbb{B}_c^n$. To avoid difficulties with optimizers, we instead supply the algorithm with a parameter vector in $\mathbb{R}^n$ that is mapped to the Poincar\'e ball using the exponential map around the origin. This is used by Lou \etal \cite{lou2020} as well. As a result, we can optimize Poincar\'e residual networks using traditional Euclidean optimizers.

A direct consequence of applying hyperbolic operations in a neural network is the large computational cost incurred by the many applications of nonlinear operations. This leads to a significant increase in memory requirements as all these intermediate steps become part of the computation graph during training. To maintain compact computation graphs, we have manually derived the backward pass of several core hyperbolic operations, namely M\"obius addition, the exponential and logarithmic maps and the conformal factor $\lambda_{\bm{x}}^c$. The use of these manually defined derivatives also reduces the size of the computation graph of many other operations defined on the Poincar\'e ball, as these generally build upon the more basic operations. We find that using manually defined derivatives reduces memory usage by approximately 30\%, but increases computation time. Due to the length of the derivations, we provide a full breakdown in the supplementary materials.

\section{Experiments}

\begin{figure*}[ht]
    \begin{subfigure}[t]{0.225\paperwidth}
        \centering
        \includegraphics[width=\textwidth]{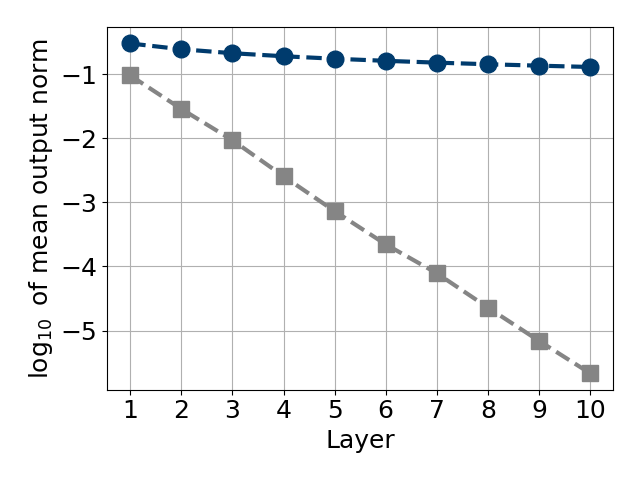}
        \caption{Output norm over layers.}
        \label{fig:init_toy_ex}
    \end{subfigure}
    \hspace{0.8em}
    \begin{subfigure}[t]{0.225\paperwidth}
        \centering
        \includegraphics[width=\textwidth]{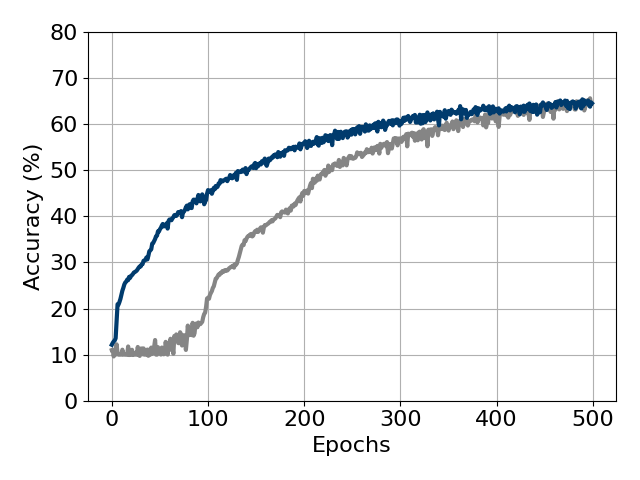}
        \caption{5-layer ConvNet training.}
        \label{fig:init_convnet_1}
    \end{subfigure}
    \hspace{0.8em}
    \begin{subfigure}[t]{0.225\paperwidth}
        \centering
        \includegraphics[width=\textwidth]{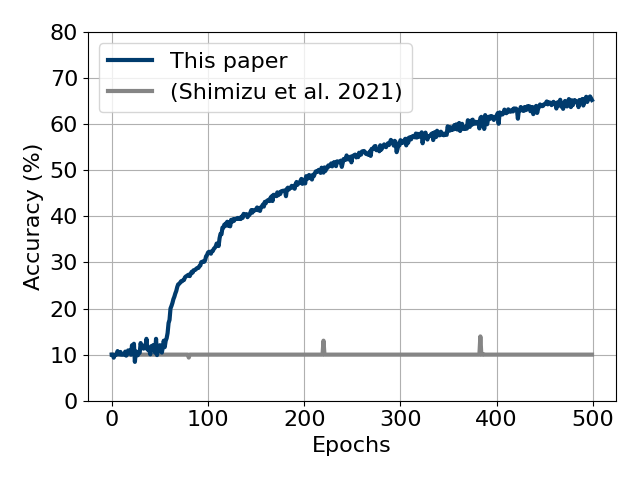}
        \caption{7-layer ConvNet training.}
        \label{fig:init_convnet_2}
    \end{subfigure}
    \centering
    \caption{\textbf{Comparison to the initialization of \cite{shimizu2021}.} In (a), we show the logarithm of the mean of the norms of each layer's output during the forward pass of an untrained 10-layer fully-connected network for random Poincar\'e gyrovectors. The figure shows that the baseline initialization is suffering from vanishing signals with outputs that collapse to the origin over multiple layers. Our identity-based initialization maintains output norms over layers. In (b) and (c), we show the test accuracy over epochs for a 5-layer and a 7-layer ConvNet. For a 5-layer network, the baseline initialization converges more slowly, while it no longer learns for 7-layers. Our initialization is preferred for training convolutional networks in the Poincar\'e ball model.}
    \label{fig:init_exps}
\end{figure*}

We investigate (i) the effect of network initialization over many layers, (ii) the effect of curvature and ReLU activations, (iii) the comparison between Fr\'echet-based and our midpoint-based batch normalization and (iv) the robustness of hyperbolic residual networks.
We seek to evaluate Poincar\'e ResNets in isolation and hence stick to minimal augmentation and fixed hyperparameters. We use random cropping and horizontal flipping with Adam optimization with fixed learning rate $10^{-3}$ and weight-decay $10^{-4}$.

\subsection{Identity initialization is norm-preserving}\label{subsect:id_init}
The approach of Shimizu \etal \cite{shimizu2021} is the current leading initialization in hyperbolic networks. This initialization, however, leads to vanishing signals, which we empirically validate here.
We take a stack of 10 Poincar\'e linear layers with a constant curvature of $c = 1$, with 20 input and output neurons.
We then perform a single forward pass on a batch of 16 Poincar\'e gyrovectors which are generated by sampling vectors in the tangent space at the origin from the multivariate normal distribution $\mathcal{N}(0, \frac{1}{10} I_{20})$ and mapping these to the Poincar\'e ball. Figure \ref{fig:init_toy_ex} shows the behaviour of the norms during the forward pass for both initialization methods. Where the baseline initialization suffers from vanishing signals, our identity initialization keeps the norms constant up to the rounding effects of the repeated application of non-linear operations.

In Figures \ref{fig:init_convnet_1} and \ref{fig:init_convnet_2} we additionally show what happens when training a simple ConvNet on CIFAR-10 with both initialization methods trained with SGD with learning rate $10^{-3}$, momentum $0.9$ and weight decay $10^{-4}$. For a 5-layer ConvNet, the baseline initialization converges more slowly. For a 7-layer ConvNet, we find that the baseline is no longer capable of learning meaningful representations. Identity-based initialization is still able to train in this setting. We conclude that our identity-based hyperbolic network initialization is preferred for training hyperbolic networks.

\subsection{Curvatures and ReLUs stabilize optimization}
\begin{figure}[t]
    \centering
    \includegraphics[width=0.47\textwidth]{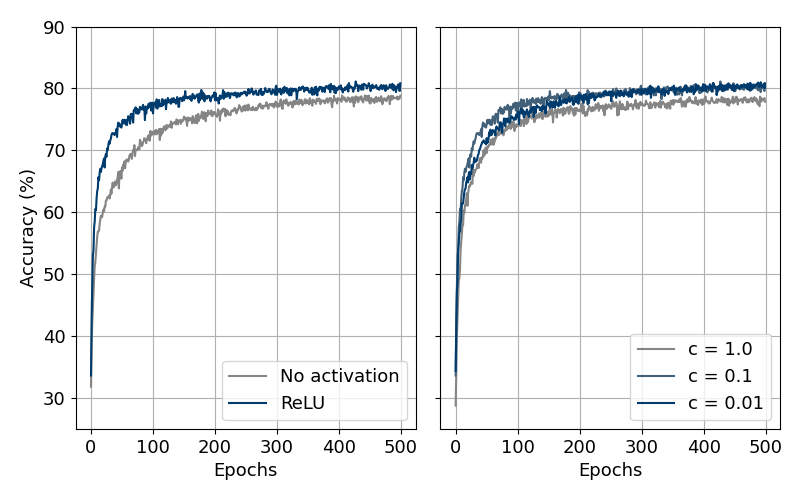}
    \caption{\textbf{ReLUs and small curvatures improve} the optimization and performance of Poincar\'e ResNet. Both experiments were performed using a small ResNet-20 with small channel widths (4, 8, 16). While hyperbolic layers are already non-linear, adding ReLUs further improves generalization. The same holds for using smaller curvatures.}
    \label{fig:curvature_and_relu_comp}
\end{figure}

Previous works claim that nonlinearities, such as the ReLU operation, are redundant in hyperbolic neural networks due to the many nonlinearities inherent to such networks \cite{ganea2018,shimizu2021}. Here, we test this claim by training a small Poincar\'e ResNet-20 on CIFAR-10 with small channel widths of (4, 8, 16) with and without the ReLU nonlinearity as activation layer. The results are shown in Figure \ref{fig:curvature_and_relu_comp} (left). We find that training with the ReLU nonlinearity leads to faster convergence and a greater final accuracy. This shows that nonlinear activation functions remain important despite the inherent nonlinearity of hyperbolic networks.

\setcounter{table}{1}
\begin{table*}
\centering
\resizebox{0.85\textwidth}{!}{
\begin{tabular}{l l cc cc cc cc cc cc}
\toprule
 & \textbf{Manifold} & \multicolumn{6}{c}{\textbf{CIFAR-10}} & \multicolumn{6}{c}{\textbf{CIFAR-100}}\\
 \cmidrule(lr){3-8} \cmidrule(lr){9-14}
& & \multicolumn{2}{c}{FPR95 ↓} & \multicolumn{2}{c}{AUROC ↑} & \multicolumn{2}{c}{AUPR ↑}  & \multicolumn{2}{c}{FPR95 ↓} & \multicolumn{2}{c}{AUROC ↑} & \multicolumn{2}{c}{AUPR ↑}\\
 \cmidrule(lr){3-4} \cmidrule(lr){5-6} \cmidrule(lr){7-8} \cmidrule(lr){9-10} \cmidrule(lr){11-12} \cmidrule(lr){13-14}
& & R20 & R32 & R20 & R32 & R20 & R32 & R20 & R32 & R20 & R32 & R20 & R32\\
\midrule
{\multirow{3}{*}{Places-365}} & Euclidean & 64.2 & 72.3 & \bf{84.7} & 82.0 & \bf{96.2} & 95.6 & 89.5 & 93.9 & 62.5 & 57.9 & 89.3 & 87.9\\
& w/ HNN++ & \bf{63.8} & 72.7 & 79.6 & 77.7 & 94.5 & 94.2 & 93.2 & 86.3 & 63.3 & 66.6 & 89.8 & 91.1\\
& \cellcolor{Gray}Poincar\'e & \cellcolor{Gray}70.2 & \cellcolor{Gray}\bf{70.7} & \cellcolor{Gray}82.3 & \cellcolor{Gray}\bf{82.6} & \cellcolor{Gray}95.7 & \cellcolor{Gray}\bf{95.9} & \cellcolor{Gray}\bf{82.8} & \cellcolor{Gray}\bf{83.8} & \cellcolor{Gray}\bf{71.5} & \cellcolor{Gray}\bf{71.1} & \cellcolor{Gray}\bf{92.3} & \cellcolor{Gray}\bf{92.2}\\
\midrule
{\multirow{3}{*}{SVHN}} & Euclidean & 97.3 & 94.7 & 68.8 & 73.4 & 92.8 & 94.1 & 99.5 & 98.8 & 43.7 & 54.6 & 83.7 & 88.2\\
& w/ HNN++ & 73.1 & 79.1 & \bf{85.5} & 82.2 & \bf{96.9} & 96.1 & 92.1 & 88.6 & 66.4 & 68.9 & 91.1 & 92.0\\
& \cellcolor{Gray}Poincar\'e & \cellcolor{Gray}\bf{66.0} & \cellcolor{Gray}\bf{69.3} & \cellcolor{Gray}85.0 & \cellcolor{Gray}\bf{83.6} & \cellcolor{Gray}96.6 & \cellcolor{Gray}\bf{96.3} & \cellcolor{Gray}\bf{76.9} & \cellcolor{Gray}\bf{83.0} & \cellcolor{Gray}\bf{76.8} & \cellcolor{Gray}\bf{72.6} & \cellcolor{Gray}\bf{94.1} & \cellcolor{Gray}\bf{92.9}\\
\midrule
{\multirow{3}{*}{Textures}} & Euclidean & 87.3 & 88.0 & 73.6 & 77.3 & 93.2 & 94.7 & 98.1 & 96.0 & 33.5 & 42.9 & 75.9 & 79.4\\
& w/ HNN++ & \bf{63.8} & \bf{56.6} & 79.6 & \bf{85.8} & 94.5 & \bf{96.6} & 85.9 & \bf{77.5} & 58.9 & 65.7 & 86.8 & 89.0\\
& \cellcolor{Gray}Poincar\'e & \cellcolor{Gray}68.2 & \cellcolor{Gray}66.2 & \cellcolor{Gray}\bf{82.1} & \cellcolor{Gray}82.3 & \cellcolor{Gray}\bf{95.5} & \cellcolor{Gray}95.6 & \cellcolor{Gray}\bf{83.9} & \cellcolor{Gray}84.2 & \cellcolor{Gray}\bf{67.7} & \cellcolor{Gray}\bf{68.8} & \cellcolor{Gray}\bf{91.0} & \cellcolor{Gray}\bf{91.5}\\
\bottomrule
\end{tabular}
}%
\caption{\textbf{Out-of-distribution detection} on CIFAR-10 and CIFAR-100 with Places365, SVHN, and DTD as out-of-distribution datasets. R20 and R32 denote ResNet-20 and ResNet-32 architectures, both with channel widths (8, 16, 32). Across different in- and out-of-distribution datasets, hyperbolic ResNets are more robust than their Euclidean counterparts.}
\label{tab:ood}
\end{table*}

Poincar\'e balls of various curvatures have similar geometric properties.
For numerical computations however, setting the right curvature impacts the down-stream performance \cite{gao2021curvature}. In this analysis, we investigate the effect of various curvatures for optimizing Poincar\'e ResNets.
We again perform the experiments on a small Poincar\'e ResNet-20 with small channel widths of (4, 8, 16) using a curvature of $1$, $0.1$ or $0.01$.
We show the results in Figure \ref{fig:curvature_and_relu_comp} (right). We first find that training with a curvature of $c=1$ leads to suboptimal accuracies. As the curvature becomes smaller, the Euclidean volume of the Poincar\'e ball increases. As a result, representing elements within this manifold using floating-point representations becomes easier with smaller curvatures. Indeed, when training with curvatures $c=0.1$ and $c=0.01$, we find that the model converges faster and has a higher final accuracy. Overall, we find that a curvature of $c=0.1$ works best for training Poincar\'e ResNets and we will use this setting for the rest of the experiments.

\setcounter{table}{0}
\begin{table}[t]
\centering
\resizebox{0.9\linewidth}{!}{
\begin{tabular}{ lc c c}
\toprule
 & & \multicolumn{1}{c}{\textbf{ResNet-20}} & \multicolumn{1}{c}{\textbf{ResNet-32}}\\
\midrule
\multirow{2}{*}{Accuracy} & Fr\'echet mean & 79.4 & 82.4 \\
& Poincar\'e midpoint & 80.9 & 81.9 \\
\midrule
Time & Fr\'echet mean & 179.0 & 169.4\\
\scriptsize{($s \text{ epoch}^{-1}$)} & Poincar\'e midpoint & 137.5 & 132.0 \\
\rowcolor{Gray} & & -23\% & -22\%\\
\bottomrule
\end{tabular}
}%
\caption{\textbf{Poincar\'e midpoints for batch normalization} in hyperbolic space are as effective for classification as Fr\'echet means while being faster to optimize.}
\label{tab:bn_results}
\end{table}

\subsection{Midpoints make batch norm efficient}
To compare the computational efficiency and the performance of our Poincar\'e midpoint batch normalization to the batch normalization by \cite{lou2020}, we perform multiple experiments using Poincar\'e ResNet-20 or Poincar\'e ResNet-32 on CIFAR-10 with small channel widths of (4, 8, 16). %
We opt for a small ResNet width and fixed hyperparameters to allow for faster evaluation, all models obtain higher scores with more depth and hyperparameter tuning.
Each network is then trained with Fr\'echet-based batch normalization \cite{lou2020} or with our Poincar\'e midpoint batch normalization.

The results of the experiment are shown in Table \ref{tab:bn_results}. First, we find that both batch normalization methods lead to similar accuracies, which indicates that Poincar\'e midpoints are as effective as Fr\'echet means for classification. Second, training a network with Poincar\'e midpoint batch normalization leads to a reduction in computation time of approximately 20-25\%. We recommend Poincar\'e midpoints when performing batch normalization in hyperbolic networks.

\begin{figure*}[ht]
    \begin{subfigure}[t]{0.25\paperwidth}
        \centering
        \includegraphics[width=\textwidth]{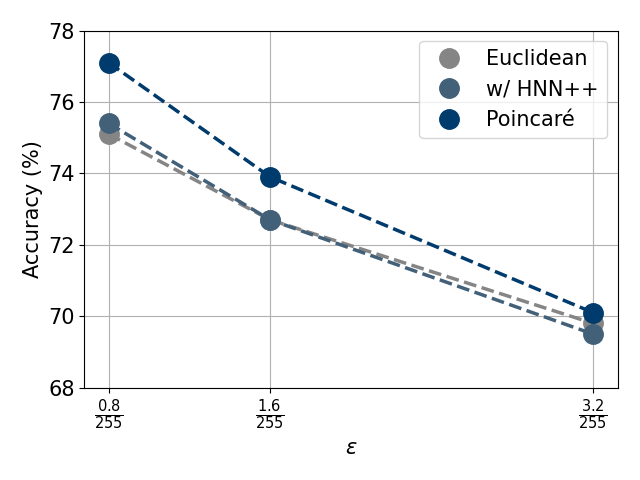}
        \caption{Adversarial robustness.}
        \label{fig:adversarial}
    \end{subfigure}
    \hspace{0.8em}
    \begin{subfigure}[t]{0.25\paperwidth}
        \centering
        \includegraphics[width=\textwidth]{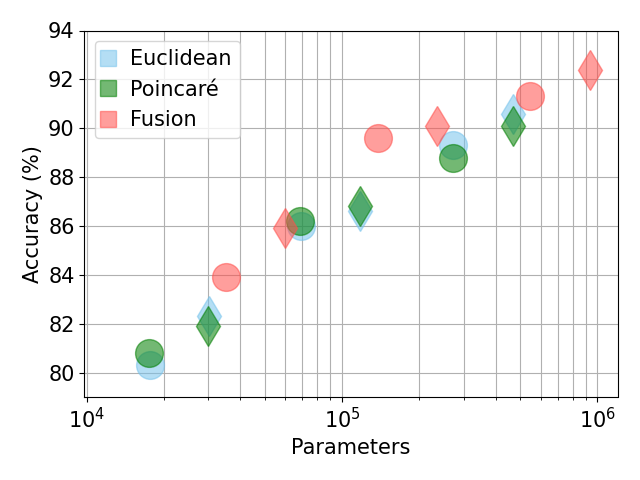}
        \caption{Hyperbolic/Euclidean fusion.}
        \label{fig:fusion}
    \end{subfigure}
    \hspace{0.8em}
    \begin{subfigure}[t]{0.25\paperwidth}
        \centering
        \vspace{-3cm}
        \includegraphics[width=\textwidth]{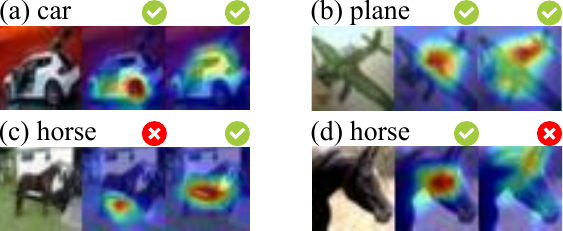}
        \vspace{.358cm}
        \caption{Grad-CAM visualizations.}
        \label{fig:gradcam}
    \end{subfigure}
    \centering
    \caption{Comparisons and fusions between hyperbolic and Euclidean ResNets. (a) \textbf{Robustness to FGSM adversarial attacks} between Euclidean and Poincar\'e ResNets. These results are obtained by attacking a {\color{black}Poincar\'e ResNet-32, or a Euclidean ResNet-32 with either a Euclidean classifier or a Poincar\'e classifier \cite{shimizu2021}}, with small channel widths (8, 16, 32), trained on CIFAR-10 to similar performance, with adversarial examples of varying perturbation sizes $\epsilon$. Poincar\'e ResNet is more robust to FGSM adversarial attacks. (b) \textbf{Fusion ResNets} plotted as a function of model parameters. The circle markers represent ResNet-20 and the diamond markers represent ResNet-32, with other differences due to varying channel widths of (4, 8, 16), (8, 16, 32) and (16, 32, 64). Fusing the Poincar\'e and Euclidean ResNets not only improves accuracy, but is more efficient than increasing the number of parameters of individual models, highlighting the strong complementary nature of learning visual representations in both spaces. (c) \textbf{Grad-CAM visualizations} of Euclidean (middle) and Poincar\'e (right) ResNets. (a) + (b) Both models predict the correct class while focusing on different discriminants in the image. (c) + (d) Failure case of respectively Euclidean and Poincar\'e ResNet due to a focus on ambiguous object parts.}
    \label{fig:robustness}
\end{figure*}

\subsection{Hyperbolic networks are robust}
Finally, we investigate the robustness and complementary nature of Poincar\'e ResNet compared to its Euclidean alternative. We investigate whether Poincar\'e ResNet is (i) robust to out-of-distribution samples, (ii) can handle adversarial examples, and (iii) learns complementary representations compared to Euclidean ResNet.

\textbf{Out-of-distribution detection.}
To check whether Poincar\'e ResNets are robust to out-of-distribution samples, we compare the out-of-distribution detection performance of Euclidean and Poincar\'e ResNet-20 and ResNet-32 with channel widths (8, 16, 32), trained on either CIFAR-10 or CIFAR-100 using the same hyperparameters and optimizer as before {\color{black}and where the Euclidean ResNets have either a Euclidean or Poincar\'e classifier \cite{shimizu2021}}. For each architecture, the Euclidean and hyperbolic variants have similar classification performance, hence any difference in out-of-distribution performance is not a result of improved training. We use the Places-365 dataset \cite{zhou2017places}, the SVHN dataset \cite{netzer2011reading} and the Textures dataset \cite{cimpoi2014describing} as out-of-distribution datasets. For detecting out-of-distribution samples, we use the energy score as introduced by Liu \etal~\cite{liu2020energy}. The comparisons are performed on the commonly used metrics FPR95, AUROC and AUPR.

The results are shown in Table \ref{tab:ood}. We find that with a ResNet-32 architecture, Poincar\'e ResNet outperforms {\color{black}both types of Euclidean ResNets on nearly all metrics for five out of six combinations of in- and out-of-distribution datasets}. With a ResNet-20, Poincar\'e ResNet is better {\color{black} for each combination with CIFAR-100 as the in-distrution dataset}. We conclude that a hyperbolic ResNet is more robust to out-of-distribution samples than its Euclidean counterpart{\color{black}, especially in the presence of many in-distribution classes}.

\textbf{Adversarial attacks.}
To see if Poincar\'e ResNet is robust to adversarial samples, we compare the performance against an adversarial attack between Euclidean ResNet-32 {\color{black}with either a Euclidean classifier or a Poincar\'e classifier \cite{shimizu2021}} and Poincar\'e ResNet-32, {\color{black}each} with channel widths (8, 16, 32), trained on CIFAR-10. Note that, after training, {\color{black}each model has} similar performance on the test set of CIFAR-10. We apply the fast gradient signed method (FGSM) \cite{goodfellow2015fgsm} attack with perturbations $\epsilon = \frac{0.8}{255}, \frac{1.6}{255}, \frac{2.4}{255}, \frac{3.2}{255}$ to the models. The results are shown in Figure \ref{fig:adversarial}. We find that Poincar\'e ResNet is more resistant to adversarial attacks than the Euclidean ResNet, even though both architectures were trained similarly and obtained similar classification performance. This result highlights the potential of hyperbolic learning in the presence of adversarial agents.
{\color{black}We note that Euclidean ResNets normally use running statistics, while Poincar\'e ResNets do not. Here we have disabled running statistics for the Euclidean models to ensure a fair comparison as running statistics make a model far more susceptible to adversarial attacks. The results when using running statistics are shown in the appendix.}

\textbf{\color{black}Complementary representations.}
To show that the representations learned by Poincar\'e ResNets are complementary to the features from Euclidean ResNets, we evaluate the performance of a fusion model, where each image is forwarded through both ResNets and the resulting logits are averaged to obtain predictions. Note that both models are trained independently and the fusion model is only evaluated with no further training being performed. The results are shown in Figure \ref{fig:fusion}. For each architecture, the performance on both manifolds is similar. Clearly, the performance of the fusion models is better than that of the individual ResNets. With respect to the number of parameters, we find that it is more efficient to create a fusion model than it is to increase the size of whichever ResNet we are using. In Figure \ref{fig:gradcam}, we also show Grad-CAM visualizations \cite{selvaraju2017grad}, highlighting that our approach focuses more on the different parts that form the object, instead of the single most discriminative component like in Euclidean ResNets.

\section{Discussion}
In this paper we propose Poincar\'e ResNet and make several contributions. First, we formulate the Poincar\'e residual block including convolutions, batchnorm, and ReLU's.
Second, we introduce an initialization that prevents vanishing signals and allows for the training of deeper models. Third, we propose a new hyperbolic batch normalization based on the Poincar\'e midpoint, which substantially increases efficiency at no cost to its performance. Fourth, we manually derive the backward pass for several operations within the Poincar\'e ball to decrease the size of the computation graphs.
Empirically, we perform initial explorations into fully hyperbolic neural networks, showing that Poincar\'e Resnets are (i) more robust to out-of-distribution samples, (ii) more robust to adversarial attacks and (iii) complementary to Euclidean networks.

\section{Acknowledgements}
Max van Spengler acknowledges the University of Amsterdam Data Science Centre for financial support.

{\small
\bibliographystyle{ieee_fullname}
\bibliography{refs}
}

\appendix
\onecolumn
\section{Gradient formulations}
\noindent
Here we provide the formulations of the manually derived gradient expressions which are used for backpropagation. For each, we provide the Jacobians ($J_\cdot$) with respect to its input variables, left-multiplied by the gradient ($u$) of its output.

\subsection{M\"obius addition}
\noindent
The M\"obius addition operation is defined as
\begin{equation}
    x \oplus_c y = \frac{(1 + 2c \langle x, y \rangle + c ||y||^2) x + (1 - c ||x||^2) y}{1 + 2c \langle x, y \rangle + c^2 ||x||^2 ||y||^2}.
\end{equation}
Its Jacobians, left-multiplied by the output gradient, can be written as
\begin{equation}\label{eq:mob_add_grad_x}
    u^T J_x (x \oplus_c y) = \frac{a}{d} u^T - \frac{2c}{d} \Big( u^T y + \frac{\theta c ||y||^2}{d} \Big) x^T + \frac{2c}{d} \Big( u^T x - \frac{\theta}{d} \Big) y^T,
\end{equation}
\begin{equation}\label{eq:mob_add_grad_y}
    u^T J_y (x \oplus_c y) = \frac{b}{d} u^T + \frac{2c}{d} \Big( u^T x - \frac{\theta}{d} \Big) x^T + \frac{2c}{d} \Big( u^T x - \frac{c ||x||^2 \theta}{d} \Big) y^T,
\end{equation}
where
\begin{equation}
    a = 1 + 2c \langle x, y \rangle + c ||y||^2,
\end{equation}
\begin{equation}
    b = 1 - c ||x||^2,
\end{equation}
\begin{equation}
    d = 1 + 2c \langle x, y \rangle + c^2 ||x||^2 ||y||^2,
\end{equation}
\begin{equation}
    \theta = a u^T x + b u^T y.
\end{equation}

\subsection{Exponential map at the origin}
\noindent
The exponential map at the origin is given by
\begin{equation}
    \exp_0^c (v) = \tanh(\sqrt{c} ||v||) \frac{v}{\sqrt{c} ||v||}.
\end{equation}
Its Jacobian, left-multiplied by the output gradient, can be written as
\begin{equation}
    u^T J_v \exp_0^c (v) = u^T v \Big( \frac{1}{||v||^2 \cosh(\sqrt{c} ||v||)^2} - \frac{\tanh(\sqrt{c} ||v||)}{\sqrt{c} ||v||^3} \Big) v^T + \frac{\tanh(\sqrt{c} ||v||)}{\sqrt{c} ||v||} u^T.
\end{equation}

\subsection{Logarithmic map at the origin}
\noindent
The logarithmic map at the origin is given by
\begin{equation}
    \log_0^c (y) = \tanh^{-1}(\sqrt{c} ||y||) \frac{y}{\sqrt{c} ||y||}.
\end{equation}
Its Jacobian, left-multiplied by the output gradient, can be written as
\begin{equation}
    u^T J_y \log_0^c (y) = u^T y \Big( \frac{1}{||y||^2 (1 - c ||y||^2)} - \frac{\tanh^{-1} (\sqrt{c} ||y||)}{\sqrt{c} ||y||^3} \Big) y^T + \frac{\tanh^{-1}(\sqrt{c} ||y||)}{\sqrt{c} ||y||} u^T.
\end{equation}

\subsection{Exponential map}
\noindent
The exponential map at $x$ is defined as 
\begin{equation}
    \exp_x^c (v) = x \oplus_c \Big(\tanh\big(\frac{\sqrt{c} \lambda_x^c ||v||}{2}\big) \frac{v}{\sqrt{c} ||v||}\Big),
\end{equation}
which we can reformulate as
\begin{equation}
    \exp_x^c (v) = x \oplus_c z_c (x, v),
\end{equation}
where
\begin{equation}
    z_c (x, v) = \tanh\big(\frac{\sqrt{c} \lambda_x^c ||v||}{2}\big) \frac{v}{\sqrt{c} ||v||}.
\end{equation}
Now we can backpropagate through this operation in two steps. First, the Jacobians of $z_c$, left-multiplied by the output gradient, can be written as
\begin{equation}
    u^T J_x z_c (x, v) = \frac{2c u^T v}{\cosh(\frac{\sqrt{c} ||v||}{1 - c ||x||^2})^2 (1 - c ||x||^2)^2} x^T,
\end{equation}
\begin{equation}
    u^T J_v z_c (x, v) = u^T v \Big( \frac{1}{||v||^2 \cosh (\frac{\sqrt{c} ||v||}{1 - c||x||^2})^2 (1 - c ||x||^2)^2} - \frac{\tanh(\frac{\sqrt{c} ||v||}{1 - c ||x||^2})}{\sqrt{c} ||v||^3} \Big) v^T + \frac{\tanh(\frac{\sqrt{c}||v||}{1 - c ||x||^2})}{\sqrt{c} ||v||} u^T.
\end{equation}
Next, for backpropagating through the M\"obius addition, we can use the expressions given in equations (\ref{eq:mob_add_grad_x}, \ref{eq:mob_add_grad_y}).

\subsection{Logarithmic map}
\noindent
The logarithmic map at $x$ is defined as
\begin{equation}
    \log_x^c (y) = \frac{2}{\sqrt{c} \lambda_x^c} \tanh^{-1} \big(\sqrt{c} ||-x \oplus_c y||\big) \frac{-x \oplus_c y}{||-x \oplus_c y||},
\end{equation}
which we can reformulate as
\begin{equation}
    \log_x^c (y) = f_c(x, z_c (x, y)) = \frac{2}{\sqrt{c} \lambda_x^c} \tanh^{-1} \big(\sqrt{c} ||z_c (x, y)||\big) \frac{z_c (x, y)}{||z_c (x, y)||},
\end{equation}
where
\begin{equation}
    z_c (x, y) = -x \oplus_c y.
\end{equation}
Again, we can backpropagate through this operation in two steps. First, we backpropagate through $z_c (x, y)$ using equations (\ref{eq:mob_add_grad_x}, \ref{eq:mob_add_grad_y}). Then, the Jacobians of $f_c (x, z)$, left-multiplied by the output gradient, can be written as
\begin{equation}
    u^T J_x f_c (x, z) = - \tanh^{-1} (\sqrt{c} ||z||) \frac{2c u^T z}{\sqrt{c} ||z||} x^T,
\end{equation}
\begin{equation}
    u^T J_z f_c (x, z) = u^T z \Big( \frac{1 - c ||x||^2}{(1 - c ||z||^2) ||z||^2} - \tanh^{-1}(\sqrt{c} ||z||) \frac{1 - c ||x||^2}{\sqrt{c} ||z||^3} \Big) z^T + \tanh^{-1}(\sqrt{c} ||z||) \frac{1 - c||x||^2}{\sqrt{c}  ||z||} u^T.
\end{equation}

\subsection{Conformal factor}
\noindent
The conformal factor is given as
\begin{equation}
    \lambda_x^c = \frac{2}{1 - c||x||^2}.
\end{equation}
Its Jacobian, multiplied by the output gradient (which is a scalar here), can be written as
\begin{equation}
    u J_x \lambda_x^c = \frac{4c u}{(1 - c||x||^2)^2} x^T.
\end{equation}

\subsection{Projection onto the Poincar\'e ball}
\noindent
An operation that is often applied in hyperbolic geometry, but rarely mentioned, is projection onto the Poincar\'e ball. This operation can be used to ensure numerical stability. It is defined as
\begin{equation}
    \text{Proj}_c (x) = x \mathbbm{1}_{\{c ||x||^2 < 1\}} (x) + \frac{x}{\sqrt{c}||x||} \mathbbm{1}_{\{c||x||^2 > 1\}} (x),
\end{equation}
where $\mathbbm{1}_A (x)$ is the indicator function, which is 1 if $x \in A$ and 0 if $x \notin A$. The Jacobian of this projection operation, left-multiplied by the output gradient, can be computed as
\begin{equation}
    u^T J_x \text{Proj}_c (x) = \Big( \mathbbm{1}_{\{c ||x||^2 < 1\}} (x) + \frac{1}{\sqrt{c} ||x||} \mathbbm{1}_{\{c||x||^2 > 1\}} (x) \Big) u^T - \Big( \frac{u^T x}{\sqrt{c} ||x||^3} \mathbbm{1}_{\{c||x||^2 > 1\}}(x) \Big) x^T.
\end{equation}

\section{Adversarial attack with running statistics}

\begin{figure}[!h]
    \centering
    \includegraphics[width=0.48\textwidth]{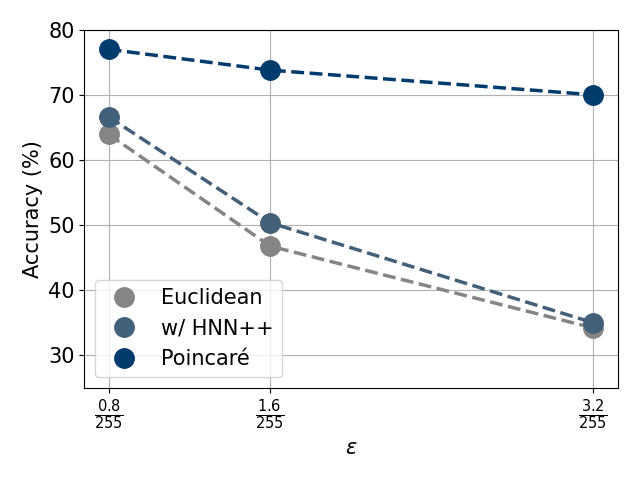}
    \caption{Adversarial attack results with running statistics for Euclidean models. The running statistics make the Euclidean models perform slightly better, but significantly more susceptible to adversarial attacks.}
    \label{fig:appendix_adversarial}
\end{figure}

\end{document}